\documentclass[runningheads]{llncs}

\usepackage[T1]{fontenc}

\usepackage{graphicx}

\usepackage{booktabs}   
\usepackage{tabularx}   
\usepackage{amsmath}
\usepackage{amssymb}
\usepackage{amsfonts}
\usepackage{threeparttable} 
\usepackage{siunitx}
\usepackage{multirow}
\usepackage{xcolor}
\usepackage{algorithm}
\usepackage{algorithmic}
\usepackage{float}
\usepackage{hyperref}

\begin{document}
\title{MMM: Quantum-Chemical Molecular Representation Learning for Combinatorial Drug Recommendation}
\titlerunning{MMM: Quantum-Chemical Molecular Representation Learning}

\author{Chongmyung Kwon\orcidID{0009-0009-0561-438X} \and
Yujin Kim\orcidID{0009-0005-3028-3165} \and
Seoeun Park\orcidID{0009-0002-0713-2746} \and
Yunji Lee\orcidID{0009-0005-9544-576X} \and
\\ Charmgil Hong\orcidID{0000-0002-8176-252X}}

\authorrunning{Kwon et al.}

\institute{Handong Global University, Pohang, South Korea\\
\email{\{21900053, 22200150, 22201019, 22100549, charmgil\}@handong.ac.kr}}

\maketitle
\begin{abstract}
Drug recommendation is an essential task in machine learning-based clinical decision support systems. However, the risk of drug-drug interactions (DDI) between co-prescribed medications remains a significant challenge. Previous studies have used graph neural networks (GNNs) to represent drug structures. Regardless, their simplified discrete forms cannot fully capture the molecular binding affinity and reactivity. Therefore, we propose Multimodal DDI Prediction with Molecular Electron Localization Function (ELF) Maps (MMM), a novel framework that integrates three-dimensional (3D) quantum-chemical information into drug representation learning. It generates 3D electron density maps using the ELF. To capture both therapeutic relevance and interaction risks, MMM combines ELF-derived features that encode global electronic properties with a bipartite graph encoder that models local substructure interactions. This design enables learning complementary characteristics of drug molecules. We evaluate MMM in the MIMIC-III dataset (250 drugs, 442 substructures), comparing it with several baseline models. In particular, a comparison with the GNN-based SafeDrug model demonstrates statistically significant improvements in the F1-score (p = 0.0387), Jaccard (p = 0.0112), and the DDI rate (p = 0.0386). These results demonstrate the potential of ELF-based 3D representations to enhance prediction accuracy and support safer combinatorial drug prescribing in clinical practice.

\keywords{Drug Recommendation \and Drug-Drug Interaction \and Multi-modality \and Electron Localization Function}
\end{abstract}

\section{Introduction}

Drug-drug interaction (DDI) refers to adverse effects that arise when multiple drugs are administered together. According to the US Food and Drug Administration (FDA)\footnote{US FDA Adverse Event Reporting System (FAERS, 2024)}, 6.3\% of the reported cases of DDI have resulted in patient mortality. This emphasizes the severity of the issue in real-world clinical settings. Prescriptions that involve potential DDI risks can lead to fatal clinical outcomes. To address this challenge, automated drug recommendation systems can be considered a promising solution to support physicians with patient-specific, data-driven, prescriptions. Such systems must simultaneously satisfy two objectives: prescribing therapeutically effective drugs and minimizing the risk of interactions among co-administered medications.

Earlier drug recommendation models, such as REverse Time AttentIoN
model (RETAIN)~\cite{choi2016retain} and DeepCare~\cite{pham2016deepcare}, utilized recurrent neural networks (RNNs)~\cite{elman1990finding} to model the longitudinal clinical trajectories observed in electronic health records (EHRs). Although effective in capturing temporal patterns, these models do not explicitly consider molecular-level drug properties, which play a crucial role in pharmacological interactions. More recent approaches, including SafeDrug~\cite{yang2021safedrug} and MoleRec~\cite{yang2023molerec}, adopted graph neural networks (GNNs)~\cite{scarselli2008graph} to represent drugs as molecular graphs. This design allows the models to learn structural features from graph-based molecular representations. However, GNN-based methods typically rely on local neighborhood aggregation, which can prevent the model from capturing global molecular properties. Furthermore, conventional graph-based representations have intrinsic limitations, as they may not adequately capture three-dimensional (3D) geometrical structures. For instance, even molecular structures that appear identical in simplified graphs can exhibit substantially different chemical reactivity and interaction profiles if their 3D configurations, such as torsion angles and bond lengths, differ~\cite{guo2022graph}. These spatial variations can affect how molecules interact with other compounds.

To overcome these limitations, it has been suggested that molecules should be represented at the quantum-chemical level or in 3D to better capture their structure~\cite{zhu2022unified,stark20223d,liu2021pre}. DDI fundamentally arises from molecular-level phenomena including steric hindrance, electronic complementarity, and localized reactive sites that affect binding affinity and metabolism~\cite{snyder2012drug,palleria2013pharmacokinetic}, all governed by the spatial distribution of electron density and molecular reactivity patterns. Among quantum-chemical descriptors~\cite{hohenberg1964inhomogeneous,politzer2002fundamental,bendova2008identifying,savin1997elf}, the Electron Localization Function (ELF) provides a continuous, 3D map of electron pair densities, specifically capturing steric hindrance regions and reactive sites that correlate with DDI occurrence mechanisms. Unlike similar quantifications such as electrostatic potential or electron density maps, ELF is particularly effective as it highlights localized bonding regions and exhibits high localization indices in covalent bond areas~\cite{becke1990simple}.

In this study, we assume that incorporating such descriptors into drug representations enables models to more accurately infer the likelihood of interactions. This inference is based on spatially diffuse and continuous electron distributions. Based on this assumption, we propose a framework that utilizes the ELF for DDI prediction, and we validate its effectiveness through extensive experiments. This proposed framework enables a richer understanding of DDI mechanisms that are often inaccessible through discrete graph-based structures.

More specifically, we propose Multimodal DDI Prediction with Molecular ELF Maps (MMM), a combinatorial drug recommendation framework that integrates longitudinal EHRs with ELF-based molecular representations. We construct ELF maps using density functional theory (DFT) ~\cite{dreizler2012density} computations and extract high-dimensional features via a pre-trained convolutional neural network (CNN)~\cite{lecun2002gradient}. These extracted molecular features are combined with patient representations to model the clinical context and drug-specific electronic behavior jointly. To the best of our knowledge, this is the first work to apply quantum-chemical molecular representations to DDI prediction. Compared to existing GNN-based methods that predict DDI through local neighborhood aggregation, MMM captures continuous electron pair localization patterns that better reflect 3D molecular reactivity relevant to DDI occurrence mechanisms. Using the MIMIC-III~\cite{johnson2016mimic} dataset, MMM introduces a modality-augmented architecture that improves predictive performance. Taken together, MMM offers a chemically and clinically more informed strategy that achieves a low DDI rate while maintaining high precision in drug recommendation.

\section{Proposed Approach}
\subsection{Preliminary}
\subsubsection{ELF-based Molecular Representation}
We construct ELF maps from simplified molecular input line entry system (SMILES)~\cite{weininger1988smiles} representations to quantize the electron density distributions of drug molecules. The process involves three main steps: first, SMILES strings are converted to optimized 3D molecular geometries using Avogadro\footnote{Avogadro version 1.2.0; developed by Hanwell, http://avogadro.cc}~\cite{hanwell2012avogadro}; second, DFT computations are performed with ORCA\footnote{ORCA version 6.0.1; developed by Neese, https://orcaforum.kofo.mpg.de}~\cite{neese2025software} at the B3LYP level of theory~\cite{tirado2008performance} to obtain electron density information. Third, ELF maps are generated using Multiwfn\footnote{Multiwfn version 3.8; developed by Lu, http://sobereva.com/multiwfn}~\cite{lu2012multiwfn,lu2024comprehensive}, with molecular planes sliced at 0.25\AA\ intervals to account for the size of the smallest (hydrogen) atoms. The resulting representations provide continuous electron localization patterns that highlight reactive sites critical for molecular interaction analysis. ELF map generation was performed for all 250 drugs using an AMD Ryzen Threadripper PRO 3955WX CPU, requiring approximately 30 hours in total. This cost is incurred only once during preprocessing, and the generated ELF maps can be stored and reused during inference.

\subsubsection{Drug Inclusion Criteria and Evaluation Strategy} 
For the purpose of DDI analysis of therapeutically prescribed medications, we exclude intravenous infusions, vitamins, and general anesthetics used for surgical purposes. In addition, we exclude nonparenteral drugs and selectively retain oral prescriptions and injectable drugs that are involved in systemic interactions. The evaluation metrics for drug prediction are calculated at levels aligned with their clinical and chemical significance. Specifically, the DDI rate is evaluated at the compound ID (CID) level~\cite{kim2025pubchem}, as the actual risks of adverse reactions are determined at the individual compound level. In contrast, the F1-score and Jaccard similarity are evaluated using the anatomical therapeutic chemical classification system, third-level codes (ATC3)~\cite{world2000collaborating} to assess therapeutic efficacy. This approach verifies clinical validity by confirming whether drugs belonging to the same therapeutic group are prescribed as alternatives.

\subsubsection{Drug Combination Recommendation}
To recommend clinically appropriate medications while minimizing the risk of DDI, we define a binary adjacency matrix $ \mathbf{D} \in \{0,1\}^{\mathcal{|M|} \times \mathcal{|M|}} $, derived from the TWOSIDES database~\cite{tatonetti2012data}. Here, $ \mathbf{D}_{ij} = 1 $ denotes a known interaction between drugs $ i $ and $ j $. Given the clinical condition of the patient at the visit $t$, the model predicts a multi-label output 
$\hat{\mathbf{m}}^{(t)} \in \{0,1\}^{|\mathcal{M}|}$, where $\hat{\mathbf{m}}_i^{(t)} = 1$ denotes that drug $ i $ is recommended.

\begin{figure}[t]
\includegraphics[width=\textwidth]{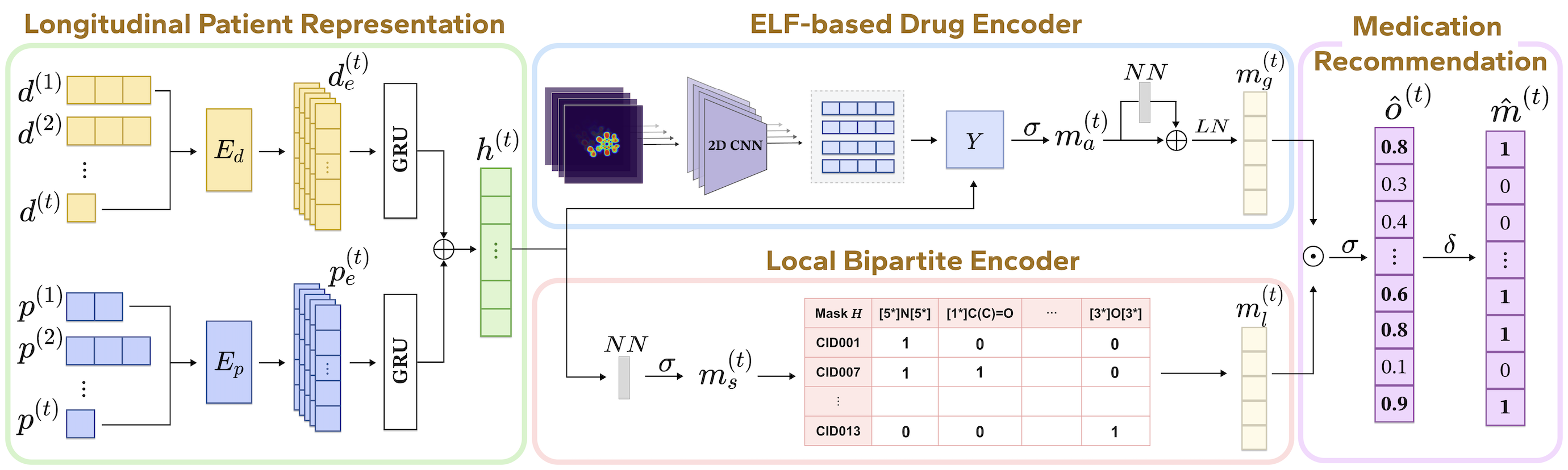}
\caption{Proposed model architecture. We first encode longitudinal EHRs into patient state vectors, which are then used to compute global and local drug representations via an ELF-based encoder and a bipartite substructure encoder. The two drug vectors are fused to generate safe and personalized drug recommendations.}
\label{fig:overview}
\end{figure}

\subsection{Model Architecture}
\subsubsection{Overview of MMM Architecture}
MMM consists of four principal components (Figure~\ref{fig:overview}):
(1) a \textbf{Longitudinal Patient Representation Module} that encodes EHRs
into patient states at each visit;
(2) an \textbf{ELF-based Drug Encoder} employs a pre-trained CNN to process 3D ELF maps, and generates global drug vectors that capture electronic interaction properties between patients and drugs. This component represents drugs through molecular electron energy density distribution and reflects how drug molecular properties influence therapeutic responses;
(3) a \textbf{Local Bipartite Encoder} infers the importance of drug substructures based on patient conditions. This module focuses on chemical substructures within drugs and learns local matching patterns between patient states and drug components;
(4) a \textbf{Medication Recommendation Module} integrates global and local drug vectors from components (2) and (3), predicts multiple drugs via a threshold-based approach. In summary, MMM represents a comprehensive model designed to derive safe and effective drug prescriptions by integrating patients temporal clinical information with both global and local molecular characteristics of drugs.

\subsubsection{Longitudinal Patient Representation} MMM takes as input three types of EHR-derived clinical information: diagnosis, procedure, and medications. Among these, the longitudinal representation of the patient used to inform medication prediction is constructed solely from the diagnosis and procedure histories. To convert EHR sequence data into uniform embedding vectors, we embed diagnosis and procedure codes from the first visit of the patient to the $t$-th visit, represented as $\mathbf{D}_e = [\mathbf{d}_e^{(1)}, \mathbf{d}_e^{(2)}, \dots, \mathbf{d}_e^{(t)}]^T,\quad \mathbf{P}_e = [\mathbf{p}_e^{(1)}, \mathbf{p}_e^{(2)}, \dots, \mathbf{p}_e^{(t)}]^T\;$. These embedded sequences are processed by an RNN model to capture temporal dependencies in patient medical histories. The RNN outputs $\mathbf{R}_{diagnosis}$ and $\mathbf{R}_{procedure}$ at the final visit are concatenated and passed through a feed-forward neural network to generate the patient representation vector $\mathbf{h}^{(t)}$. We denote by $\mathbf{W}_1$ the weight matrix used in the feed-forward neural network.
\begin{equation}
\mathbf{h}^{(t)} = \mathbf{NN}(\mathbf{R}_{diagnosis} \; \| \; \mathbf{R}_{procedure}; \:\mathbf{W}_1)
\label{eq:patient_representation}
\end{equation}

\subsubsection{ELF-based Drug Encoder}
To enable accurate patient–drug matching, it is essential to capture molecular-level interactions that govern the efficacy of the drug. Each drug $i$ in vocabulary $\mathcal{M}$ is represented through its ELF map, which encodes spatial electron distributions. We extract image patches $\mathbf{I}_i = \{ \mathbf{I}_i^{(1)}, \mathbf{I}_i^{(2)}, \dots \}$ from these maps to model local electronic patterns. A pre-trained CNN model processes each patch $\mathbf{I}_i^{(j)}$ to extract feature vectors, capturing drug-specific electronic behavior through quantum-chemical molecular representations. This operation aggregates features across all patches to produce a unified drug representation $\mathbf{C}_i$ that preserves quantum-chemical information. The resulting drug matrix $\mathbf{C} \in \mathbb{R}^{|\mathcal{M}| \times \text{dim}}$ contains embeddings for all drugs. 

To enable direct comparison with the patient representation vector $\mathbf{{h}}^{(t)}$, we align dimensions through a multi-layer perceptron (MLP)~\cite{rumelhart1986learning}. This produces final drug embeddings $\mathbf{Y}$ that preserve molecular specificity while being dimensionally compatible to compute patient-drug similarity.

\begin{equation}
\mathbf{Y = f_{\text{MLP}}(C)} \in \mathbb{R}^{|\mathcal{M}| \times \text{dim}}
\end{equation}
To select appropriate drugs for the current clinical condition of a patient, it is essential to have a mechanism that quantifies the interaction between patient representations and drug characteristics. For this purpose, we perform a dot-product operation between the patient representation vector $\mathbf{h}^{(t)}$ at time point $t$ and the ELF-based drug embedding matrix $\mathbf{Y}$. This mechanism aligns the patient state with drug representations, enabling the model to identify which drugs are most relevant for the given patient context. Through this operation, we derive a suitability score vector $\mathbf{m}_a^{(t)}$ that quantifies how well each drug corresponds to the current patient state. This score serves as a selection criterion for drug recommendation and forms the basis for computing the global drug vector $\mathbf{m}_g^{(t)}$. Attention-like suitability scores $\mathbf{m}_a^{(t)}$ capture nuanced relationships between patient characteristics and drug properties, providing a principled approach to drug selection that goes beyond simple similarity matching. We apply a feed-forward neural network with residual connections and layer normalization (LN)~\cite{ba2016layer} to transform $\mathbf{m}_a^{(t)}$ into the final global drug vector $\mathbf{m}_g^{(t)}$, which stabilizes training and preserves gradient flow for optimization. $\mathbf{W}_2$ denotes the weights of this feed-forward neural network. The global drug vector $\mathbf{m}_g^{(t)}$ represents the final drug output score from the ELF-based drug encoder, providing essential information for patient-specific drug recommendation.

\begin{equation}
{\mathbf{m}_a^{(t)} = \sigma\left(\mathbf{h}^{(t)} \cdot \mathbf{Y}^T\right)}, \quad
{{\mathbf{m}}_g^{(t)} = \mathrm{\mathbf{LN}}\left({\mathbf{m}}_a^{(t)}+ \mathrm{\mathbf{NN}}({\mathbf{m}}_a^{(t)}; \mathbf{W}_2)\right)}
\label{eq:global_attention}
\end{equation}

\subsubsection{Local Bipartite Encoder}
Recognizing that the relevance of drug substructures depends on patient conditions, we utilize a patient-specific local drug representation~\cite{yang2021safedrug}.
We first take the patient representation vector $\mathbf{h}^{(t)}$ at time $t$ as input and generate a vector $\mathbf{m}_s^{(t)} \in \mathbb{R}^{|\mathcal{S}|}$ through a learnable function, which represents the importance of each substructure across the entire substructure space $\mathcal{S}$. This vector quantifies which substructures are more therapeutically relevant for the current state of the patient. 

To model the relationship between drugs and their constituent substructures, we apply the breaking retrosynthetically interesting chemical substructures (BRICS) decomposition algorithm~\cite{degen2008art} to segment each drug into a set of substructures. Based on this decomposition, we construct a binary mask matrix $\mathbf{H}$ that encodes the inclusion relationships between drugs and substructures, where $\mathbf{H}_{i,j}=1$ indicates that drug $i$ contains substructure $j$.

Finally, we transform the original drug parameter weights $\mathbf{W}_3$ by element-wise multiplication with the mask matrix $\mathbf{H}$ to obtain substructure-level weights, and perform matrix multiplication with the patient-based substructure importance vector $\mathbf{m}_s^{(t)}$. This process aggregates the substructures of each drug that are relevant to the specific patient, thereby producing a patient-specific local drug vector $\mathbf{m}_l^{(t)}$. Through this approach, identical drugs can have different importance scores depending on patient conditions. This enables personalized drug representations that better reflect individual therapeutic needs.

\begin{equation}
{{\mathbf{m}}_l^{(t)} = \mathrm{\mathbf{NN}} \left( {\mathbf{m}}_s^{(t)}; \, \mathbf{W}_3 \odot \mathbf{H} \right)}
\label{eq:local_drug}
\end{equation}

\subsubsection{Medication Representation}
The final medication probability vector ${\mathbf{\hat{o}}^{(t)}}$ is computed through the sigmoid function applied to the element-wise product of the ELF-based representation $\mathbf{m}_g^{(t)}$ and the local bipartite representation $\mathbf{m}_l^{(t)}$.

\begin{equation}
{\mathbf{\hat{o}}^{(t)} = \sigma\left(\mathbf{m}_g^{(t)} \odot \mathbf{m}_l^{(t)} \right)}
\label{eq:medication_rep}
\end{equation}

\begin{algorithm}[H]
\caption{One Training Epoch of the MMM}
\label{tab:Algorithm}
\small
\begin{algorithmic}[1]
\REQUIRE Training data $\mathcal{T}$, CNN embeddings $\mathbf{C}$, mask matrix $\mathbf{H}$ 
\ENSURE Updated model parameters $\theta$
\FOR{each patient $j$ in $\mathcal{T}$}
    \STATE Select patient $j$'s EHR sequence $\mathbf{X}_j$
    \FOR{$t = 1$ to $|\mathbf{X}_j|$}
        \STATE Encode the longitudinal patient history via RNN to obtain $\mathbf{h}^{(t)}$ \hfill (Eq.~1)
        \STATE Extract ELF-based global drug embeddings $\mathbf{C}_i$ via CNN and max pooling
        \STATE Project $\mathbf{C}_i$ through MLP to obtain the embeddings $\mathbf{Y}$ \hfill (Eq.~2)
        \STATE Compute $\mathbf{m}_a^{(t)}$ via dot-product between $\mathbf{h}^{(t)}$ and $\mathbf{Y}^T$ \hfill (Eq.~3)
        \STATE Apply $LN$ to $\mathbf{m}_a^{(t)}$ to obtain the global drug vector $\mathbf{m}_g^{(t)}$ \hfill (Eq.~3)
        \STATE Compute the substructure importance vector $\mathbf{m}_s^{(t)}$
        \STATE Derive the local drug vector $\mathbf{m}_l^{(t)}$ using the bipartite mask $\mathbf{H}$ \hfill (Eq.~4)
        \STATE Perform an element-wise product of $\mathbf{m}_g^{(t)}$ and $\mathbf{m}_l^{(t)}$ to predict the multi-hot medication vector ${\mathbf{\hat{o}}^{(t)}}$ \hfill (Eq.~5)
        \STATE Compute $L_{bce}$, $L_{multi}$, and $L_{DDI}$
        \STATE Compute total loss 
            $L = \beta \cdot (\alpha \cdot L_{bce} + (1 - \alpha) \cdot L_{multi}) + (1 - \beta) \cdot L_{DDI}$ 
        \STATE Update $\theta$ using Adam optimizer (lr=5e-5)
    \ENDFOR
\ENDFOR
\end{algorithmic}
\end{algorithm}

\subsubsection{Model Training and Loss Function} Algorithm~\ref{tab:Algorithm} describes the one training epoch of MMM. The medication recommendation task is formulated as a multi-label binary classification problem, where MMM independently predicts the prescription probability for each drug in the medication vocabulary $\mathcal{M}$. 

For model optimization, MMM employs a multiobjective framework that combines three loss components: binary cross-entropy loss $\mathcal{L}_{\mathrm{bce}}$, multilabel margin loss $\mathcal{L}_{\mathrm{multi}}$, and DDI-aware loss $\mathcal{L}_{\mathrm{DDI}}$. Unlike prior methods that rely on dynamically adjusting loss weights based on DDI rate, MMM uses fixed hyperparameters $\alpha$ and $\beta$ to ensure stable and consistent training. The model is optimized using the Adam optimizer with a learning rate of 5e-5, and the final checkpoint is selected based on the lowest DDI rate achieved across epochs.

\section{Experiments and Results}

We conduct comparative experiments against existing baselines and perform ablation studies to quantitatively evaluate both DDI reduction and prediction performance of the proposed model, MMM. To further assess practical clinical applicability, we conduct case studies to examine whether the model recommends prescriptions comparable to those actually prescribed to patients. The entire dataset was partitioned into training, validation and test sets in a ratio of $\frac{2}{3} : \frac{1}{6} : \frac{1}{6}$, with stratified holdout sampling applied to mitigate class imbalance and ensure consistent evaluation. For the ELF-based drug encoder, a pre-trained EfficientNetV2-L model~\cite{tan2021efficientnetv2} is employed to generate ELF embedding vectors.

\subsubsection{Dataset and Metrics} The experiments were conducted using the MIMIC-III dataset~\cite{johnson2016mimic}, with summary statistics of the preprocessed data presented in Table~\ref{tab:DataStatistics}. To evaluate the performance of medication recommendation, we employed four metrics: DDI rate, Jaccard similarity, F1-score, and the average number of prescribed medications.  

\begin{table}[t]
\small
\caption{Data Statistics. (D: Diagnosis, M: Medication, P: Procedure)}
\label{tab:DataStatistics}
\centering
\begin{tabular}{l|l||l|l}
\toprule
{\bfseries Items} & {\bfseries Size} & {\bfseries Items} & {\bfseries Size} \\
\midrule
\# of visits/\# of patients & 14,057 / 5,413 & avg./max \# of visits & 2.60 / 29 \\
D. / P. / M. space size & 1,942 / 1,399 / 250 & avg./max \# of D. per visit & 10.38 / 128 \\
total \# of DDI pairs & 4,918 & avg./max \# of P. per visit & 3.85 / 50 \\
total \# of substructures & 442 & avg./max \# of M. per visit & 7.67 / 68 \\
\bottomrule
\end{tabular}
\end{table}

\begin{table}[t]
\small
\caption{Performance Comparison on MIMIC-III (recorded DDI rate is 0.2509)}
\centering
\label{tab:PerformanceComparison}
\begin{tabular}{
    l
    |>{\centering\arraybackslash}p{2.4cm}
    |>{\centering\arraybackslash}p{2.4cm}
    |>{\centering\arraybackslash}p{2.4cm}
    |>{\centering\arraybackslash}p{2.4cm}
}
\toprule
\textbf{Model} & \textbf{DDI Rate} & \textbf{Jaccard} & \textbf{F1-score} & \textbf{Avg. \# Drugs} \\
\midrule
Random Forest &
\(0.3652 \pm 0.0018\) &
\(0.3123 \pm 0.0019\) &
\(0.4628 \pm 0.0023\) &
\(4.8476 \pm 0.0113\) \\
RETAIN &
\(0.3325 \pm 0.0098\) &
\(0.4882 \pm 0.0129\) &
\(0.6319 \pm 0.0114\) &
\(5.7883 \pm 0.1757\) \\
MoleRec &
\(0.0760 \pm 0.0031\) &
\(0.7384 \pm 0.0127\) &
\(0.8353 \pm 0.0094\) &
\(14.9414 \pm 1.1696\) \\
SafeDrug &
\(0.0742 \pm 0.0026\) &
\(0.7488 \pm 0.0081\) &
\(0.8434 \pm 0.0064\) &
\(13.4697 \pm 1.4838\) \\
\midrule
\textbf{MMM} &
\(0.0673 \pm 0.0049^{*}\) &
\(0.7608 \pm 0.0066^{*}\) &
\(0.8498 \pm 0.0046^{*}\) &
\(12.5239 \pm 0.9008\) \\
\bottomrule
\end{tabular}
\begin{tablenotes}
\small
\item $^*$ : Statistical significance of MMM over the best baseline, under the paired $t$-test (DDI, Jaccard, and F1).
\end{tablenotes}
\end{table}

\subsection{Quantitative Evaluation}

Table~\ref{tab:PerformanceComparison} shows the performance comparison between MMM and four baselines: Random Forest~\cite{breiman2001random}, RETAIN~\cite{choi2016retain}, MoleRec~\cite{yang2023molerec}, and SafeDrug~\cite{yang2021safedrug}. These baselines were selected to cover various methodological approaches: Specifically, Random Forest serves as a fundamental machine learning benchmark, while RETAIN captures temporal  patterns in EHRs. In contrast, MoleRec and SafeDrug represent advanced GNN-based approaches for molecular representation learning. All experiments were repeated 10 times with bootstrapped sampling, and statistical significance was evaluated using two-sided paired $t$-tests.

The results reveal distinct performance patterns across different methodological approaches. In terms of DDI rate, Random Forest and RETAIN exhibit the highest DDI rate of 0.3652 and 0.3325, respectively as they do not incorporate DDI and molecular structure information in their predictions. In contrast, MoleRec and SafeDrug achieve relatively lower DDI rates of 0.0760 and 0.0742, respectively by leveraging molecular graph representations, suggesting that molecular encoders contribute to improved prescription safety. MMM recorded the lowest DDI rate of 0.0673 by modeling DDIs using 3D ELF-based molecular energy maps. This demonstrates that our approach can capture molecular interactions more effectively than conventional molecular graph methods.

Furthermore, MMM also demonstrated improvements in therapeutic prediction objectives. In terms of predictive performance metrics, it achieved a Jaccard similarity of 0.7608 and an F1-score of 0.8498, outperforming all baseline models. These results indicate that the proposed method can deliver robust predictive performance across both DDI reduction and therapeutic recommendation tasks.

\subsection{Ablation Study}

\begin{table}
\small
\caption{Ablation Study: Effect of Each Component on Model Performance}\label{tab1}
\centering
\label{tab:AblationStudy}
\begin{tabular}{
    >{\raggedright\arraybackslash}p{2.3cm}
    |>{\centering\arraybackslash}p{2.3cm}
    |>{\centering\arraybackslash}p{2.3cm}
    |>{\centering\arraybackslash}p{2.3cm}
    |>{\centering\arraybackslash}p{2.3cm}
}
\toprule
\textbf{Model} & \textbf{DDI Rate} & \textbf{Jaccard} & \textbf{F1-score} & \textbf{Avg. \# Drugs} \\
\midrule
\textbf{w/o Bipartite Encoder} & 
\(0.0776 \pm 0.0023\) & 
\(0.7450 \pm 0.0132\) & 
\(0.8363 \pm 0.0104\) & 
\(15.2948 \pm 1.0907\) \\
\textbf{w/o ELF Encoder} & 
\(0.0610 \pm 0.0068\) & 
\(0.7182 \pm 0.0297\) & 
\(0.8195 \pm 0.0231\) & 
\(15.2336 \pm 1.8888\) \\
\textbf{MMM} & 
\(0.0673 \pm 0.0049\) & 
\(0.7608 \pm 0.0066\) & 
\(0.8498 \pm 0.0046\) & 
\(12.5239 \pm 0.9008\) \\ 
\bottomrule
\end{tabular}
\end{table}

The ablation study in Table~\ref{tab:AblationStudy} was designed to validate our assumption that ELF-based representations and bipartite graph representations play complementary roles in drug recommendation.
When using only the ELF encoder, the DDI rate increased to 0.0776, which is 15.3\% higher than the complete model. Interestingly, the metrics of therapeutic effectiveness remained at high levels, with an F1-score of 0.8363 and a Jaccard similarity of 0.7450. However, it showed a tendency to prescribe an average of 15.29 drugs, which confirms that without molecular substructure information, interaction risks increase due to over-prescription. This result supports that substructure-level information is crucial for DDI prediction. Conversely, when only the bipartite encoder was used, the F1-score dropped to 0.8195, and the Jaccard decreased significantly to 0.7182, while this configuration showed the lowest DDI rate at 0.0610. Despite prescribing an average of 15.23 drugs, the low DDI rate can be attributed to the bipartite encoder focusing on DDI avoidance, leading to a tendency to select prescription combinations that insufficiently reflect therapeutic similarity in the absence of entire molecular representations.

On the other hand, the complete MMM achieved a DDI rate of 0.0673. It maintains a low DDI rate while recording the highest performance in metrics of therapeutic effectiveness. This demonstrates that the ELF encoder provides information on molecular electronic structure for therapeutic effectiveness, while the bipartite encoder provides substructure patterns to avoid DDI, working in a complementary manner. These results of the ablation study support our core design decisions of combining 3D quantum-chemical representations with substructure-based analysis.

\subsection{Case Study}

\begin{table}[t]
\small
\caption{Case Study: Patient from MIMIC-III with multiple diagnoses}
\label{tab:CaseStudy}
\renewcommand{\arraystretch}{1.3}
\setlength{\tabcolsep}{4pt}
\centering
\begin{tabularx}{\textwidth}{|l|l|X|}
\hline
\multicolumn{2}{|c|}{} & \textbf{Patient 1} \\
\hline
\multicolumn{2}{|l|}{\textbf{Diagnosis}} & Morbid obesity, Hypertension, Osteoarthrosis, Disorders of circulatory system, Accidental hemorrhage \\
\hline
\multicolumn{2}{|l|}{\textbf{Prescribed Medications}} & \textcolor{red}{Gabapentin}, Warfarin, Argatroban, \textcolor{red}{Midazolam}, Cefazolin, \textcolor{red}{Pantoprazole}, \textcolor{red}{Metoprolol}, \textcolor{red}{Furosemide} \\
\hline
\multirow{3}{*}{\shortstack[l]{\textbf{Recommended}\\\textbf{Medications}}}
 & \textbf{SafeDrug} & Bisacodyl, Docusate, \textcolor{red}{Acetaminophen}, Hydromorphone, \textcolor{red}{Metoprolol}, Warfarin, \textcolor{red}{Pantoprazole}, Lisinopril, Morphine, Oxycodone \\
\cline{2-3}
 & \textbf{MoleRec} & \textcolor{red}{Acetaminophen}, Bisacodyl, \textcolor{red}{Furosemide}, Docusate, Hydromorphone, \textcolor{red}{Pantoprazole}, Lisinopril, Warfarin, Morphine \\
\cline{2-3}
 & \textbf{MMM} & \textcolor{red}{Acetaminophen}, Bisacodyl, Docusate, Hydromorphone, \textcolor{red}{Metoprolol}, \textcolor{red}{Pantoprazole}, Clopidogrel, Lisinopril, \textcolor{red}{Ondansetron}, Morphine, Oxycodone, Famotidine \\
\hline

\end{tabularx}
\begin{tablenotes}
\small
\item The recorded prescriptions in the dataset resulted in a DDI rate of 0.3214, whereas SafeDrug, MoleRec, and MMM achieved lower DDI rates of 0.0833, 0.0909, and 0.0667, respectively. Red color indicates interacting medications. 
\end{tablenotes}
\end{table}

Table~\ref{tab:CaseStudy} summarizes an analysis of prescription safety for a patient with multiple diagnoses, including morbid obesity, hypertension, osteoarthritis, circulatory disorders, and hemorrhage. In the dataset prescriptions, several high-risk drug combinations were identified, including Warfarin–Argatroban, which is contraindicated due to bleeding risk, and Gabapentin–Midazolam, associated with central nervous system depression. While both SafeDrug and MoleRec successfully removed some of these high-risk agents, they retained combinations such as Warfarin co-prescribed with opioid analgesics, which are known to further increase bleeding and sedation risk. In contrast, MMM successfully avoided all major high-risk DDIs and recommended safer alternatives, preserving therapeutic effectiveness while effectively reducing the DDI rate. This outcome suggests that the proposed model can simultaneously achieve diagnosis-aware clinical safety and proactive interaction avoidance.

\section{Conclusion and Discussion}
In this work, we introduced MMM, a multimodal drug recommendation framework that integrates longitudinal EHR data with ELF-based quantum-chemical molecular representations to address drug–drug interaction risks. 
Our model outperformed existing graph-based baselines on the MIMIC-III dataset in terms of both DDI rate reduction and recommendation accuracy. 
Unlike prior methods relying on topological graphs, MMM utilizes continuous 3D electron localization maps computed via DFT. This enables it to encode spatially diffuse electronic features relevant to intermolecular interactions. 
As a result, the model can account for physicochemical properties such as bonding character and reactive site localization, which are not readily captured by conventional representations.

Several limitations remain. The current framework operates under a binary classification setup that treats all DDIs as equally undesirable. 
This abstraction limits its alignment with real-world prescribing decisions, where interaction risks must be balanced with therapeutic priorities. 
Future work will expand the descriptor space to incorporate more expressive molecular and clinical features, and assess their task-specific utility. We aim to integrate mechanistic DDI classifications as well as continuous severity scores to improve interpretability and clinical relevance. Although this study focuses on 250 commonly prescribed drugs from the MIMIC-III dataset to reflect real-world prescription patterns, we plan to extend our evaluation to a larger set of drugs and additional datasets to assess generalizability of our proposed framework across diverse clinical settings.

\section*{Acknowledgements}
This research was supported by the MSIT (Ministry of Science and ICT), Korea, under (1) the Global Research Support Program in the Digital Field (RS-2024-00431394) and (2) the National Program for Excellence in SW (2023-0-00055) supervised by the IITP (Institute for Information \& Communications Technology Planning \& Evaluation).

%
%
%
%
%
\bibliographystyle{unsrt}
\bibliography{references}

\begin{thebibliography}{10}

\bibitem{choi2016retain}
Edward Choi, Mohammad~Taha Bahadori, Jimeng Sun, Joshua Kulas, Andy Schuetz, and Walter Stewart.
\newblock Retain: An interpretable predictive model for healthcare using reverse time attention mechanism.
\newblock {\em Advances in neural information processing systems}, 29, 2016.

\bibitem{pham2016deepcare}
Trang Pham, Truyen Tran, Dinh Phung, and Svetha Venkatesh.
\newblock Deepcare: A deep dynamic memory model for predictive medicine.
\newblock In {\em Advances in Knowledge Discovery and Data Mining: 20th Pacific-Asia Conference, PAKDD 2016, Auckland, New Zealand, April 19-22, 2016, Proceedings, Part II 20}, pages 30--41. Springer, 2016.

\bibitem{elman1990finding}
Jeffrey~L Elman.
\newblock Finding structure in time.
\newblock {\em Cognitive science}, 14(2):179--211, 1990.

\bibitem{yang2021safedrug}
Chaoqi Yang, Cao Xiao, Fenglong Ma, Lucas Glass, and Jimeng Sun.
\newblock Safedrug: Dual molecular graph encoders for recommending effective and safe drug combinations.
\newblock {\em arXiv preprint arXiv:2105.02711}, 2021.

\bibitem{yang2023molerec}
Nianzu Yang, Kaipeng Zeng, Qitian Wu, and Junchi Yan.
\newblock Molerec: Combinatorial drug recommendation with substructure-aware molecular representation learning.
\newblock In {\em Proceedings of the ACM web conference 2023}, pages 4075--4085, 2023.

\bibitem{scarselli2008graph}
Franco Scarselli, Marco Gori, Ah~Chung Tsoi, Markus Hagenbuchner, and Gabriele Monfardini.
\newblock The graph neural network model.
\newblock {\em IEEE transactions on neural networks}, 20(1):61--80, 2008.

\bibitem{guo2022graph}
Zhichun Guo, Kehan Guo, Bozhao Nan, Yijun Tian, Roshni~G Iyer, Yihong Ma, Olaf Wiest, Xiangliang Zhang, Wei Wang, Chuxu Zhang, et~al.
\newblock Graph-based molecular representation learning.
\newblock {\em arXiv preprint arXiv:2207.04869}, 2022.

\bibitem{zhu2022unified}
Jinhua Zhu, Yingce Xia, Lijun Wu, Shufang Xie, Tao Qin, Wengang Zhou, Houqiang Li, and Tie-Yan Liu.
\newblock Unified 2d and 3d pre-training of molecular representations.
\newblock In {\em Proceedings of the 28th ACM SIGKDD conference on knowledge discovery and data mining}, pages 2626--2636, 2022.

\bibitem{stark20223d}
Hannes St{\"a}rk, Dominique Beaini, Gabriele Corso, Prudencio Tossou, Christian Dallago, Stephan G{\"u}nnemann, and Pietro Li{\`o}.
\newblock 3d infomax improves gnns for molecular property prediction.
\newblock In {\em International Conference on Machine Learning}, pages 20479--20502. PMLR, 2022.

\bibitem{liu2021pre}
Shengchao Liu, Hanchen Wang, Weiyang Liu, Joan Lasenby, Hongyu Guo, and Jian Tang.
\newblock Pre-training molecular graph representation with 3d geometry.
\newblock {\em arXiv preprint arXiv:2110.07728}, 2021.

\bibitem{snyder2012drug}
Ben~D Snyder, Thomas~M Polasek, and Matthew~P Doogue.
\newblock Drug interactions: principles and practice.
\newblock {\em Australian prescriber}, 35(3), 2012.

\bibitem{palleria2013pharmacokinetic}
Caterina Palleria, Antonello Di~Paolo, Chiara Giofr{\`e}, Chiara Caglioti, Giacomo Leuzzi, Antonio Siniscalchi, Giovambattista De~Sarro, and Luca Gallelli.
\newblock Pharmacokinetic drug-drug interaction and their implication in clinical management.
\newblock {\em Journal of research in medical sciences: the official journal of Isfahan University of Medical Sciences}, 18(7):601, 2013.

\bibitem{hohenberg1964inhomogeneous}
Pierre Hohenberg and Walter Kohn.
\newblock Inhomogeneous electron gas.
\newblock {\em Physical review}, 136(3B):B864, 1964.

\bibitem{politzer2002fundamental}
Peter Politzer and Jane~S Murray.
\newblock The fundamental nature and role of the electrostatic potential in atoms and molecules.
\newblock {\em Theoretical Chemistry Accounts}, 108(3):134--142, 2002.

\bibitem{bendova2008identifying}
Lada Bendov{\'a}-Biedermannov{\'a}, Pavel Hobza, and Ji{\v{r}}{\'\i} Vondr{\'a}{\v{s}}ek.
\newblock Identifying stabilizing key residues in proteins using interresidue interaction energy matrix.
\newblock {\em Proteins: Structure, Function, and Bioinformatics}, 72(1):402--413, 2008.

\bibitem{savin1997elf}
Andreas Savin, Reinhard Nesper, Steffen Wengert, and Thomas~F F{\"a}ssler.
\newblock Elf: The electron localization function.
\newblock {\em Angewandte Chemie International Edition in English}, 36(17):1808--1832, 1997.

\bibitem{becke1990simple}
Axel~D Becke and Kenneth~E Edgecombe.
\newblock A simple measure of electron localization in atomic and molecular systems.
\newblock {\em The Journal of chemical physics}, 92(9):5397--5403, 1990.

\bibitem{dreizler2012density}
Reiner~M Dreizler and Eberhard~KU Gross.
\newblock {\em Density functional theory: an approach to the quantum many-body problem}.
\newblock Springer Science \& Business Media, 2012.

\bibitem{lecun2002gradient}
Yann LeCun, L{\'e}on Bottou, Yoshua Bengio, and Patrick Haffner.
\newblock Gradient-based learning applied to document recognition.
\newblock {\em Proceedings of the IEEE}, 86(11):2278--2324, 2002.

\bibitem{johnson2016mimic}
Alistair~EW Johnson, Tom~J Pollard, Lu~Shen, Li-wei~H Lehman, Mengling Feng, Mohammad Ghassemi, Benjamin Moody, Peter Szolovits, Leo Anthony~Celi, and Roger~G Mark.
\newblock Mimic-iii, a freely accessible critical care database.
\newblock {\em Scientific data}, 3(1):1--9, 2016.

\bibitem{weininger1988smiles}
David Weininger.
\newblock Smiles, a chemical language and information system. 1. introduction to methodology and encoding rules.
\newblock {\em Journal of chemical information and computer sciences}, 28(1):31--36, 1988.

\bibitem{hanwell2012avogadro}
Marcus~D Hanwell, Donald~E Curtis, David~C Lonie, Tim Vandermeersch, Eva Zurek, and Geoffrey~R Hutchison.
\newblock Avogadro: an advanced semantic chemical editor, visualization, and analysis platform.
\newblock {\em Journal of cheminformatics}, 4:1--17, 2012.

\bibitem{neese2025software}
Frank Neese.
\newblock Software update: The orca program system—version 6.0.
\newblock {\em Wiley Interdisciplinary Reviews: Computational Molecular Science}, 15(2):e70019, 2025.

\bibitem{tirado2008performance}
Julian Tirado-Rives and William~L Jorgensen.
\newblock Performance of b3lyp density functional methods for a large set of organic molecules.
\newblock {\em Journal of chemical theory and computation}, 4(2):297--306, 2008.

\bibitem{lu2012multiwfn}
Tian Lu and Feiwu Chen.
\newblock Multiwfn: A multifunctional wavefunction analyzer.
\newblock {\em Journal of computational chemistry}, 33(5):580--592, 2012.

\bibitem{lu2024comprehensive}
Tian Lu.
\newblock A comprehensive electron wavefunction analysis toolbox for chemists, multiwfn.
\newblock {\em The Journal of Chemical Physics}, 161(8), 2024.

\bibitem{kim2025pubchem}
Sunghwan Kim, Jie Chen, Tiejun Cheng, Asta Gindulyte, Jia He, Siqian He, Qingliang Li, Benjamin~A Shoemaker, Paul~A Thiessen, Bo~Yu, et~al.
\newblock Pubchem 2025 update.
\newblock {\em Nucleic Acids Research}, 53(D1):D1516--D1525, 2025.

\bibitem{world2000collaborating}
World~Health Organization et~al.
\newblock Collaborating centre for drug statistics methodology, guidelines for atc classification and ddd assignment.
\newblock {\em WHO Collaborating Centre for Drug Statistics Methodology}, 18, 2000.

\bibitem{tatonetti2012data}
Nicholas~P Tatonetti, Patrick~P Ye, Roxana Daneshjou, and Russ~B Altman.
\newblock Data-driven prediction of drug effects and interactions.
\newblock {\em Science translational medicine}, 4(125):125ra31--125ra31, 2012.

\bibitem{rumelhart1986learning}
David~E Rumelhart, Geoffrey~E Hinton, and Ronald~J Williams.
\newblock Learning representations by back-propagating errors.
\newblock {\em nature}, 323(6088):533--536, 1986.

\bibitem{ba2016layer}
Jimmy~Lei Ba, Jamie~Ryan Kiros, and Geoffrey~E Hinton.
\newblock Layer normalization.
\newblock {\em arXiv preprint arXiv:1607.06450}, 2016.

\bibitem{degen2008art}
Jorg Degen, Christof Wegscheid-Gerlach, Andrea Zaliani, and Matthias Rarey.
\newblock On the art of compiling and using'drug-like'chemical fragment spaces.
\newblock {\em ChemMedChem}, 3(10):1503, 2008.

\bibitem{tan2021efficientnetv2}
Mingxing Tan and Quoc Le.
\newblock Efficientnetv2: Smaller models and faster training.
\newblock In {\em International conference on machine learning}, pages 10096--10106. PMLR, 2021.

\bibitem{breiman2001random}
Leo Breiman.
\newblock Random forests.
\newblock {\em Machine learning}, 45:5--32, 2001.

\end{thebibliography}
\end{document}